# Tighter Variational Representations of $f$-Divergences via Restriction to Probability Measures

**Avraham Ruderman**                                          AVRAHAM.RUDERMAN@NICTA.COM.AU
**Mark D. Reid**                                              MARK.REID@ANU.EDU.AU
**Darío García-García**                                       DARIO.GARCIA@ANU.EDU.AU
Research School of Computer Science, Australian National University and NICTA, Canberra, Australia

**James Petterson**                                          JAMES.PETTERSON@NICTA.COM.AU
NICTA, Canberra, Australia

## Abstract

We show that the variational representations for $f$-divergences currently used in the literature can be tightened. This has implications to a number of methods recently proposed based on this representation. As an example application we use our tighter representation to derive a general $f$-divergence estimator based on two i.i.d. samples and derive the dual program for this estimator that performs well empirically. We also point out a connection between our estimator and MMD.

## 1. Introduction

An important class of discrepancy measures between probability distributions is the family of $f$-divergences (also known as Ali-Silvey (Ali & Silvey, 1966) or Csisz·r divergences (Csiszár, 1967)). These include the variational divergence, the Hellinger distance, and the well-known Kullback-Leibler (KL) divergence. Estimates of these measures from two i.i.d. samples can be used to test whether or not those samples come from similar distributions. Due to the convexity of the eponymous $f$ function defining them, $f$-divergences can be expressed variationally as the maximum of an optimisation problem. This variational representation has been recently used for $f$-divergence estimation (Nguyen et al., 2010; 2007; Kanamori et al., 2011) homogeneity testing (Kanamori et al., 2011) and parameter estimation (Broniatowski & Keziou, 2009).

This paper stems from a simple observation: the vari-ational representation currently being used in the literature fails to take into account the fact that divergences are defined between *probability* distributions. In an analysis similar to that of Altun and Smola's (Altun & Smola, 2006), we present a derivation of a quantifiably tighter variational representation of $f$-divergences. Our derivation *restricts* the convex dual used in the variational representation of an $f$-divergence to the space of probability distributions. This generalises similar observations for the specific case of the KL divergence, such as Banerjee's compression lemma (Banerjee, 2006). These results are given in the remainder of this section.

Clearly, use of this tighter variational representation in any algorithm relying on the previous bounds would be advantageous. Thus, we suspect this tighter bound will find broad applicability as the weaker variational form is already widely used. As an example of this, in section 2. we derive a general dual program for an RKHS estimator which has a simple closed form for any $f$-divergence. We show this dual program has a natural interpretation as a trade off between the minimisation of the $f$-divergence in question and the minimum mean discrepancy (MMD) (Gretton et al., 2008) between the empirical distributions. Experiments in section 3 confirm that the tighter variational form underlying our approach leads to better KL divergence estimates than the well known estimator of (Nguyen et al., 2010; 2007) which is based on the looser representation. We also provide an empirical comparison to other state of the art methods for $f$-divergence estimation.

### 1.1. Convex Duality

We briefly introduce some key ideas from convex duality. The reader is referred to (Barbu & Precupanu,

1986) for details. For functions $f : X \to \mathbb{R}$ defined over a Banach space $X$ the *(Fenchel or convex) dual* $f^\star : X^\star \to \mathbb{R}$ of a function is defined over the dual space $X^\star$ by $f^\star(x^\star) := \sup_{x \in X} \langle x^\star, x \rangle - f(x)$ where $\langle \cdot, \cdot \rangle$ is the dual pairing of $X$ and $X^\star$. In finite dimensional spaces such as $\mathbb{R}^d$ the dual space is also $\mathbb{R}^d$ and $\langle \cdot, \cdot \rangle$ is the usual inner product. The *bidual* $f^{\star\star} : X \to \mathbb{R}$ of $f$ is just the dual of $f^\star$ (restricted to $X$), that is, $f^{\star\star}(x) = \sup_{x^\star \in X^\star} \langle x, x^\star \rangle - f^\star(x^\star)$. For convex, lower semi-continuous (l.s.c.) functions $f$ the bidual is the identity transformation, that is, $f^{\star\star} = f$. As seen below, this fact forms the basis of many variational representations of operators defined by convex functions.[1]

We make use of a few specific, one-dimensional instances of duals. Specifically, when $f(t) = |t - 1|$ we have $f^\star(t^\star) = t^\star$ for $t^\star \in [-1, 1]$; and when $f(t) = -\ln t$ for $t > 0$ we have $f^\star(t^\star) = -1 - \ln t^\star$ for $t^\star < 0$. Further details and properties of convex duals can be found in texts on convex analysis, e.g., (Hiriart-Urruty & Lemaréchal, 2001).

One obvious but important property we make use of is that the restriction of a supremum leads to smaller optima. Specifically, if $X' \subseteq X$ then $\sup_{x \in X'} \phi(x) \le \sup_{x \in X} \phi(x)$. When applied to convex duals, this means if $R \subseteq X^\star$ is a restriction of the dual space then

$$f(x) \ge f^R(x) := \sup_{x^\star \in R} \langle x^\star, x \rangle - f^\star(x^\star) \quad \text{for all } x \in X. \tag{1}$$

## 1.2. Variational Approximations of $f$-divergences

An $f$-divergence is defined via a convex function $f : [0, \infty) \to \mathbb{R}$ satisfying $f(1) = 0$. Given such a function, the *$f$-divergence from a finite measure $P$ to a distribution $Q$* defined on a common space $X$ is defined[2] as

$$\mathbb{I}_f(P, Q) := \mathbb{E}_Q \left[ f\left( \frac{dP}{dQ} \right) \right] = \int_X f\left( \frac{dP}{dQ}(x) \right) dQ(x)$$

if $P \ll Q$ and $+\infty$ otherwise. We will refer to the definition above as the *general (or unrestricted) $f$-divergence* in contrast to the *restricted $f$-divergence* that is only defined when $P$ and $Q$ are *both* probability distributions. When necessary, the restricted

$f$-divergence will be distinguished by a superscript $R$: $\mathbb{I}_f^R(P, Q)$. We emphasise this distinction to later show how a tighter variational representation can be obtained from explicitly taking into account the restriction. Several common divergences are members of this class: the variational divergence is obtained by choosing $f(t) = |t - 1|$, Hellinger divergence via $f(t) = \sqrt{t^2 - 1}$, and the KL divergence via $f(t) = t \ln t$ (see, e.g., (Reid & Williamson, 2011)). For technical reasons we also require $f$ to be lower semi-continuous. All $f$-divergences discussed above and used in practice satisfy this condition.

As in (Altun & Smola, 2006; Barbu & Precupanu, 1986; Broniatowski & Keziou, 2009), we now wish to consider $f$-divergences as acting over spaces of functions. Given a measure $\mu$ over $X$ (with some $\sigma$-algebra), the norms $\|g\|_1 := \int_X |g| \, d\mu$ and $\|g\|_\infty := \inf \{K \ge 0 : |g(x)| \le K \text{ for } \mu\text{-almost all } x\}$ can be used to define the space of absolutely integrable functions $L^1(\mu) := \{g : X \to \mathbb{R} : \|g\|_1 < \infty\}$ and its dual space $L^\infty(\mu) := \{g : X \to \mathbb{R} : \|g\|_\infty < \infty\}$ of functions with bounded essential supremum. The space of probability densities w.r.t. $\mu$ will be denoted $\Delta(\mu) := \{g \in L^1(\mu) : g \ge 0, \|g\|_1 = 1\}$. On finite domains $X = \{x_1, \ldots, x_n\}$, the space of densities will be denoted $\Delta^n$. A *general $f$-divergence* can be seen as acting on $L^1(Q)$ by defining $\mathbb{I}_{f,Q}(r) := \mathbb{E}_Q[f(r)]$ for all $r \in L^1(Q)$. The *restricted $f$-divergence* is then just $\mathbb{I}_{f,Q}^R(r) := \mathbb{I}_{f,Q}(r)$ when $r \in \Delta(Q)$ and $+\infty$ otherwise.

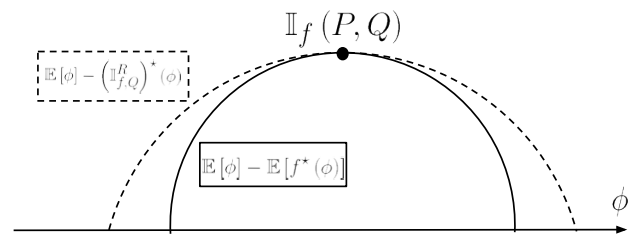

*Figure 1.* Illustration of Theorem 1. The dashed line and solid line represent our new expression and the expression used by (Nguyen et al., 2010) respectively as they vary over $L^\infty(Q)$. While both expressions have the same supremum, everywhere else ours is closer to the supremum.

As shown below, these functions are convex and lower semi-continuous and therefore admit dual representations. As explored in (Nguyen et al., 2010; Altun & Smola, 2006) and in section 2 below, variational representations such as these readily admit approximation techniques via a restriction of the optimisation for $\mathbb{I}_{f,Q}$ to functions from $\mathcal{F} \subseteq L^{\infty}(Q)$ (e.g., an RKHS). Our main result is that the restricted variational form guarantees tighter lower bounds on $f$-divergences than the unrestricted form for any choice of function class.

---

[1] For finite dimensional spaces, this is, in some sense, the *only* dual that can be used for the kind of variational representations we are interested in (see (Artstein-Avidan & Milman, 2009) for details).

[2] The choice of order of the arguments $P$ and $Q$ is arbitrary and other authors, notably (Nguyen et al., 2010), define $f$-divergence in terms of $dQ/dP$.

**Theorem 1.** *For any distributions $P$ and $Q$ such that $P \ll Q$ we have*

$$\mathbb{I}_f(P,Q) = \sup_{\phi \in L^{\infty}(Q)} \mathbb{E}_P[\phi] - \left(\mathbb{I}_{f,Q}^R\right)^{\star}(\phi) \qquad (2)$$

*and for any choice of function class $\mathcal{F} \subseteq L^{\infty}(Q)$*

$$\mathbb{I}_f(P,Q) \geq \sup_{\phi \in \mathcal{F}} \mathbb{E}_P[\phi] - \left(\mathbb{I}_{f,Q}^R\right)^{\star}(\phi)$$
$$\geq \sup_{\phi \in \mathcal{F}} \mathbb{E}_P[\phi] - \mathbb{E}_Q[f^{\star}(\phi)]. \qquad (3)$$

We illustrate the result of Theorem 1 in Figure 1. While our new expression and that of (Nguyen et al., 2010) have the same supremum, our expression is closer to the supremum at every point $\phi \in L^{\infty}(Q)$. In particular, if one restricts $\phi$ to vary over a subset $\mathcal{F} \subset L^{\infty}$ then the supremum of our expression over $\mathcal{F}$ will yield a better estimate of $\mathbb{I}_f(P,Q)$.

*Proof.* We first establish that $\mathbb{I}_{f,Q}$ is a convex function over $L^1(Q)$ via Proposition 2.7 in (Barbu & Precupanu, 1986) which states that for any finite measure $Q$ over $X$ and any proper convex l.s.c. function $f : [0,\infty) \to \mathbb{R} \cup \{+\infty\}$ the function $F : L^q(Q) \to (-\infty, +\infty]$ is convex and l.s.c when defined by $F(u) := \int_X f(u)\, dQ$ for $f(u) \in L^1(Q)$ and $+\infty$ otherwise. Since $\mathbb{I}_{f,Q}(r) = \int_X f(r)\, dQ$ and $f$ satisfies the conditions of the proposition, we have that $\mathbb{I}_{f,Q}$ is convex and l.s.c.

The variational representation for $\mathbb{I}_{f,Q}$ is then obtained by using Lemma 4.5.8 of (Dembo & Zeitouni, 2009). This states that for any Banach space $X$ with dual $X^{\star}$, if $F : X \to \mathbb{R} \cup \{+\infty\}$ is convex and l.s.c. then $F(x) = \sup_{x^{\star} \in X^{\star}} \{\langle x^{\star}, x \rangle - F^{\star}(x^{\star})\}$. Since $\mathbb{I}_{f,Q}$ is convex and l.s.c. on $L^1(Q)$ we see that

$$\mathbb{I}_{f,Q}(r) = \sup_{\phi \in L^{\infty}(Q)} \left\{\mathbb{E}_Q[\phi r] - (\mathbb{I}_{f,Q})^{\star}(\phi)\right\} = (\mathbb{I}_{f,Q})^{\star\star}(r)$$

for all $r \in L^1(Q)$ since $L^{\infty}(Q)$ is the dual space to $L^1(Q)$. In particular, we can conclude that for all $r \in L^1(Q)$ and all $\phi \in L^{\infty}(Q)$

$$\mathbb{I}_{f,Q}(r) \geq \mathbb{E}_Q[\phi r] - (\mathbb{I}_{f,Q})^{\star}(\phi). \qquad (4)$$

Now, since $\mathbb{I}_{f,Q}^R$ is defined to be $\mathbb{I}_{f,Q}$ on $\Delta(Q)$ and $+\infty$ on the rest of $L_1(Q)$ we see that for all $\phi \in L^{\infty}(Q)$

$$\left(\mathbb{I}_{f,Q}^R\right)^{\star}(\phi) = \sup_{r \in L^1(Q)} \left\{\mathbb{E}_Q[\phi r] - \mathbb{I}_{f,Q}^R(r)\right\}$$
$$= \sup_{p \in \Delta(Q)} \left\{\mathbb{E}_Q[\phi p] - \mathbb{I}_{f,Q}(p)\right\}. \qquad (5)$$

This implies that for all $\phi \in L^{\infty}(Q)$ and for all $p \in \Delta(Q)$ we have $\left(\mathbb{I}_{f,Q}^R\right)^{\star}(\phi) + \mathbb{I}_{f,Q}(p) \geq \mathbb{E}_Q[\phi p]$ which rearranged and optimised over $\phi \in L^{\infty}(Q)$ yields (2). We now observe that (5) is just a constrained version of the optimisation defining $(\mathbb{I}_{f,Q})^{\star}$, and so by (1) we must have $\left(\mathbb{I}_{f,Q}^R\right)^{\star} \leq (\mathbb{I}_{f,Q})^{\star}$. Substituting this into (4) we see that for any $p \in \Delta(Q), \phi \in L^{\infty}(Q)$

$$\mathbb{I}_{f,Q}(p) \geq \mathbb{E}_Q[\phi p] - \left(\mathbb{I}_{f,Q}^R\right)^{\star}(\phi) \geq \mathbb{E}_Q[\phi p] - (\mathbb{I}_{f,Q})^{\star}(\phi).$$

Taking supremums over these inequalities for $\phi \in \mathcal{F} \subseteq L^{\infty}(Q)$ yields (3), as required. $\qquad \square$

For the particular case of KL divergence estimation the above theorem specialises to show that for any choice of $\phi \in L^{\infty}(Q)$

$$KL(P,Q) \geq \mathbb{E}_P[\phi] - \ln \mathbb{E}_Q\left[e^{\phi}\right] \geq \mathbb{E}_P[\phi] - \mathbb{E}_Q\left[e^{\phi} + 1\right].$$

The first inequality is the well-known representation of the KL divergence in the large deviations literature (Donsker & Varadhan, 1983) which has been rediscovered in the PAC-Bayes community as the compression lemma (Banerjee, 2006). Although the second inequality can be obtained immediately from the fact that $-\log(y) \geq -y + 1$ for all $y > 0$, Theorem (1) shows that a similar result holds for general $f$-divergences.

## 2. Estimation using RKHS methods

Given two samples[3] $X_n := \{x_1, \dots, x_n\}$ and $Y_n := \{y_1, \dots, y_n\}$ from $P$ and $Q$ respectively, we wish to use the empirical measures $P_n := \frac{1}{n}\sum_{i=1}^n \delta_{x_i}$ and $Q_n := \frac{1}{n}\sum_{i=1}^n \delta_{y_i}$ as proxies for $P$ and $Q$. However, since we no longer have $P_n \ll Q_n$ some form of smoothing is required. We make use of the restricted dual variational representation of the $f$-divergence and choose a sufficiently constrained function class over which the supremum is taken when computing the dual. Specifically, we let $\mathcal{H} \subset L_{\infty}(Q)$ be an RKHS with reproducing kernel $K$ and corresponding feature map $\Phi$ and choose some convex regulariser $\Omega : \mathcal{H} \to \mathbb{R} \cup \{+\infty\}$, for example, some function of the RKHS norm $\|\cdot\|_{\mathcal{H}}$. As in (Nguyen et al., 2010), our estimator is then defined via the dual representation of $\mathbb{I}_{f,Q_n}^R$ that takes

---

[3] We assume that the samples are of equal size for simplicity. The analysis also goes through in the general case.

into account the regulariser $\Omega$, i.e.,

$$E(P_n, Q_n) := \sup_{h \in \mathcal{H}} \left\{ \mathbb{E}_{P_n}[h] - \left(\mathbb{I}_{f,Q_n}^R\right)^\star(h) - \Omega(h) \right\}.$$
(6)

The following result gives an explicit dual optimisation program for computing $E$ in the RKHS $\mathcal{H}$.

**Theorem 2.** *Let $\mathcal{H}$ be an RKHS of functions over $X$ with associated feature map $\Phi$. Then the estimator $E(P_n, Q_n)$ satisfies, for all $P_n$ and $Q_n$*

$$E(P_n, Q_n) = \min_{\alpha \in \Delta^n} \left\{ \frac{1}{n} \sum_{i=1}^n f(n\alpha_i) \right.$$
$$\left. + \Omega^\star \left( \frac{1}{n} \sum_{i=1}^n \Phi(x_i) - \frac{1}{n} \sum_{i=1}^n n\alpha_i \Phi(y_i) \right) \right\}$$
(7)

*where the minimisation is over the $n$-simplex $\Delta^n$.*

*Proof.* The proof techniques here are based on those in (Nguyen et al., 2007). Since $\mathcal{H}$ is a RKHS, we can represent each function $h \in \mathcal{H}$ by $h(x) = \langle w, \Phi(x) \rangle$ for $x \in X$ where $\Phi$ is the feature map corresponding to $K$. In this case, the estimator in (6) is given by

$$\sup_w \left\{ \frac{1}{n} \sum_{i=1}^n \langle w, \Phi(x_i) \rangle - \left(\mathbb{I}_{f,Q_n}^R\right)^\star(\langle w, \Phi(\cdot) \rangle) - \Omega(w) \right\}.$$

Letting $\psi(w) := -\frac{1}{n} \sum_{i=1}^n \langle w, \Phi(x_i) \rangle$ and $\varphi(w) := \left(\mathbb{I}_{f,Q_n}^R\right)^\star(\langle w, \Phi(\cdot) \rangle)$ and substituting into the above expression gives

$$\sup_w \left\{ \langle w, 0 \rangle - (\psi(w) + \varphi(w) + \Omega(w)) \right\} = (\psi + \varphi + \Omega)^\star(0)$$
(0)

by the definition of a dual. By the infimal convolution theorem (Rockafellar, 1997) we therefore have

$$E(P_n, Q_n) = \min_{s,r} \left\{ \psi^\star(s) + \varphi^\star(r) + \Omega^\star(-s-r) \right\}.$$
(8)

Now, since $\psi$ is linear in $w$ its dual is simply

$$\psi^\star(s) = \begin{cases} 0 & \text{, if } s = -\frac{1}{n} \sum_{i=1}^n \Phi(x_i) \\ +\infty & \text{, otherwise.} \end{cases}$$

To compute the dual of $\varphi$ we observe that $\left(\mathbb{I}_{f,Q_n}^R\right)^\star(h)$ only depends on the values of $h$ at $y_1, \dots, y_n$ so we can

write

$$\varphi^\star(r) = \sup_w \left\{ \langle w, r \rangle - \left(\mathbb{I}_{f,Q_n}^R\right)^\star(\langle w, \Phi(\cdot) \rangle) \right\}$$
$$= \sup_{w,\alpha,h} \left\{ \langle w, r \rangle - \left(\mathbb{I}_{f,Q_n}^R\right)^\star(h) \right.$$
$$\left. - \sum_{i=1}^n \alpha(y_i)\left(\langle w, \Phi(y_i) \rangle - h(y_i)\right) \right\}$$
$$= \sup_{w,\alpha,h} \left\{ \langle w, r \rangle - \left(\mathbb{I}_{f,Q_n}^R\right)^\star(h) \right.$$
$$\left. - \left\langle w, \sum_{i=1}^n \alpha(y_i)\Phi(y_i) \right\rangle + \frac{1}{n} \sum_{i=1}^n n\,\alpha(y_i)\,h(y_i) \right\}$$

by the introduction of Lagrange multipliers $\alpha(y_i)$ for the constraints $h(y_i) = \langle w, \Phi(y_i) \rangle$. Noting that $\frac{1}{n} \sum_{i=1}^n n\alpha(y_i)h(y_i) = \mathbb{E}_{Q_n}[n\alpha h]$ we get

$$\varphi^\star(r) = \sup_{w,\alpha,h} \left\{ \mathbb{E}_{Q_n}[n\alpha h] - \left(\mathbb{I}_{f,Q_n}^R\right)^\star(h) \right.$$
$$\left. - \left\langle w, r - \sum_{i=1}^n \alpha_i \Phi(y_i) \right\rangle \right\}$$
$$= \sup_\alpha \left\{ \sup_h \left\{ \mathbb{E}_{Q_n}[n\alpha h] - \left(\mathbb{I}_{f,Q_n}^R\right)^\star(h) \right\} \right.$$
$$\left. + \sup_w \left\{ \left\langle w, \sum_i \alpha_i \Phi(y_i) - r \right\rangle \right\} \right\}$$
$$= \sup_\alpha \left\{ \mathbb{I}_{f,Q_n}^R(n\alpha) : \sum_i \alpha_i \Phi(y_i) = r \right\}$$

since the first inner supremum is the bidual of $\mathbb{I}_{f,Q_n}^R$ and the second supplies the constraint. Furthermore, since $\Phi(y_i)$ are linearly independent, each $r$ uniquely determines $\alpha$ at $y_1, \dots, y_n$ so $\varphi^\star(r) = \mathbb{I}_{f,Q_n}^R(n\alpha)$. Substituting $\psi^\star$, $\varphi^\star$, and the corresponding constraints on $s$ and $r$ back into the minimisation (8) and noting that $\mathbb{I}_{f,Q_n}^R$ is $+\infty$ for $n\alpha \notin \Delta(Q_n) \simeq \Delta^n$ gives the required result:

$$E(P_n, Q_n) = \min_{\alpha \in \Delta^n} \left\{ \frac{1}{n} \sum_{i=1}^n f(n\alpha_i) \right.$$
$$\left. + \Omega^\star \left( \frac{1}{n} \sum_{i=1}^n \Phi(x_i) - \frac{1}{n} \sum_{i=1}^n (n\alpha_i)\Phi(y_i) \right) \right\}.$$
$\square$

We note that $E(P_n, Q_n)$ is not a direct estimate of $\mathbb{I}_f(P, Q)$ due to the inclusion of the regularisation term $\Omega(h)$. However, $\varphi^\star(r) = \mathbb{I}_{f,Q_n}^R(n\alpha) = \frac{1}{n} \sum_{i=1}^n f(n\alpha_i)$ can be used as an empirical estimate of $\mathbb{I}_f(P, Q)$ once the values of $\alpha_i$ are obtained and $\hat{r} = n\alpha$ can be seen as an estimate of $dP/dQ$. Thus, the theorem above gives an easily implementable algorithm for estimating $f$-divergences.

Computing $\Omega^\star$ for particular choices of $\Omega$ in equation (7) immediately gives the following corollary which defines two concrete estimators.

**Corollary 3.** *For* $\Omega(g) = \frac{\lambda_n}{2}\|g\|_{\mathcal{H}}^2$

$$E(P_n, Q_n) = \min_{\alpha \in \Delta^n} \left\{ \frac{1}{n} \sum_{i=1}^{n} f\left(n\alpha_i\right) \right.$$
$$\left. + \frac{1}{2\lambda_n} \left\| \frac{1}{n} \sum_{i=1}^{n} \Phi\left(x_i\right) - \frac{1}{n} \sum_{i=1}^{n} n\alpha_i \Phi\left(y_i\right) \right\|_{\mathcal{H}}^2 \right\} \tag{9}$$

*and for* $\Omega(g) = \sqrt{\lambda_n}\|g\|_{\mathcal{H}}$

$$E(P_n, Q_n) = \min_{\alpha \in \Delta^n} \left\{ \frac{1}{n} \sum_{i=1}^{n} f\left(n\alpha_i\right) : \right.$$
$$\left. \left\| \frac{1}{n} \sum_{i=1}^{n} \Phi\left(x_i\right) - \frac{1}{n} \sum_{i=1}^{n} n\alpha_i \Phi\left(y_i\right) \right\|_{\mathcal{H}} \leq \sqrt{\lambda_n} \right\}. \tag{10}$$

The minimisation in (10) is similar to the one discussed in (Altun & Smola, 2006) which is concerned with density estimation from a single sample. Our estimator can be seen as an extension of that procedure to the two sample setting. The estimator M2 proposed in (Nguyen et al., 2010) uses square norm regularisation in a two sample setting and is therefore directly comparable to (9). The key difference is that we restrict the minimisation to $\Delta^n$ whereas the M2 minimisation is over $\alpha \geq 0$.

### 2.1. Connections with Maximum Mean Discrepancy

The relation of the optimisation program in (9) to the original $f$-divergence is compelling. The first term $\frac{1}{n}\sum_{i=1}^{n} f\left(n\alpha_i\right)$ is simply the empirical estimate of the $f$ divergence of the likelihood ratio $\frac{dP}{dQ}$ since each $n\alpha_i = \frac{\alpha_i}{1/n}$ is an estimate of $\frac{dP}{dQ}(y_i)$ when $Q_n$ is taken to be uniform over $y_1, \ldots, y_n$. Since $f(1) = 0$, minimising this term alone would force the $\alpha_i = \frac{1}{n}$ for all $i$. In this sense, the first term can be seen as a kind of generalised MaxEnt regularisation.

The second term can be seen as a term that forces the $\alpha_i$ terms to "match" empirical means of the feature vectors $\Phi(x_i)$ and $\Phi(y_i)$. Following (Gretton et al., 2008), we can formalise the observation regarding the second term by considering the mean map $\mu$ from distributions $R$ over $X$ to functions in an RKHS $\mathcal{H}$ defined by $R \mapsto \mu[R] := \mathbb{E}_{x \sim R}[\Phi(x)]$. The maximum mean discrepancy (MMD) between distributions $P$ and $Q$ is then defined to be the distance between their respective images under $\mu$, that is, $\text{MMD}(P, Q) = \|\mu[P] - \mu[Q]\|_{\mathcal{H}}$. The second term in our estimator is then just $\text{MMD}^2(P_n, \alpha)$ which is a measure of the discrepancy between the distributions corresponding to the densities $dP_n$ and $n\alpha dQ_n$. Thus, minimising that term alone corresponds to an unregularised estimation of the density ratio $dP/dQ$. Similarly, for other choices of regularisation $\Omega$ which are a function of $\|\cdot\|_{\mathcal{H}}$, this "data matching" term will be a dual function of $\text{MMD}(P_n, \alpha)$.

This analysis also leads to an intuitive explanation why we should use the regularisation schedule $\lambda_n = \Theta\left(n^{-1}\right)$ as per (Nguyen et al., 2009). It was shown in (Gretton et al., 2008) that the MMD estimator

$$\sqrt{n} \left\| \frac{1}{n} \sum_{i=1}^{n} \Phi\left(y_i\right) - \frac{1}{n} \sum_{i=1}^{n} \Phi\left(x_i\right) \right\|_{\mathcal{H}}$$

converges to a normal distribution with constant variance. If $\alpha$ is suitably bounded away from infinity, the same holds for the second term in (9) as long as $\lambda_n = \Theta\left(n^{-1}\right)$. If $\lambda_n$ is of smaller order, then the MMD term will eventually dominate the general MaxEnt term which converges to a positive constant if $P \neq Q$. On the other hand if $\lambda_n$ diminishes more slowly, then the MMD term will go to zero even for an incorrect density ratio.

## 3. Experiments

Theorem 1 shows the restricted variational bound derived here is strictly tighter than the one proposed by (Nguyen et al., 2010) (henceforth NWJ) for every function $r \in L^1(Q)$ except when $r = dP/dQ$ in which case they coincide and attain the optimum. This suggests that the optimisation problem derived using our tighter bound should result in an estimator with a smaller bias. This section presents some empirical results demonstrating this improvement. We also conducted experiments to compare our method and Nguyen et al's method to others methods in the literature which are not based on variational representations of $f$-divergence. While these non variational methods are not the focus of the experiments, we include them here as they may be of interest to others. In the context of the current work however, we emphasise the superior performance of our estimator compared to that of NWJ illustrating the utility of the tighter variational representation. We include the following recent estimators for comparison:

- Wang et al (Wang et al., 2009): This estimator is based on nearest neighbour estimates of the two densities and does not make use of a variational representation.

- Kanamori et al (Kanamori et al., 2009): This is a least-squares estimator for the density ratio, bypassing individual density estimations. Once the density ratio is estimated, it can be directly plugged in the $f$-divergence formulae. We also experimented with another density ratio estimation method (Sugiyama et al., 2008), with very similar results.

- García et al (García-García et al., 2011): This estimator uses nearest neighbour misclassification rates and a reformulation of $f$-divergences in terms of risks.

### 3.1. Method

Both our estimator based on (9) and the M2 estimator of NWJ were implemented using the nonlinear convex optimisation routine from the python package CVX-OPT to perform the optimization. The implementation of the Wang et al. (Wang et al., 2009) estimator (henceforth WKV) was based on the cKDTree nearest neighbour routine from the SciPy library. Kanamori et al [4]. (uLSIF) and García et al [5] ($(f, l)$) algorithms were implemented using code provided by the respective authors [6]. The method for choosing the parameters $\lambda_n$ and $\sigma$ for the NWJ estimator are not specified in (Nguyen et al., 2010). For both NWJ and our estimator, we therefore set $\lambda_n = \frac{1}{n}$ (as discussed above) and set $\sigma$ to the sample variance over $X_n \cup Y_n$ to ensure invariance with respect to rescaling of the data.

In every experiment, the distributions $P$ and $Q$ were set to beta distributions $B(\alpha, \beta)$ for some choice of parameters $\alpha, \beta > 0$. Beta distributions were chosen as they cover a wide variety of shapes and have a KL divergence with the following analytic form

$$KL(B(\alpha_1, \beta_1), B(\alpha_2, \beta_2)) = \ln \frac{B(\alpha_2, \beta_2)}{B(\alpha_1, \beta_1)}$$

$$-d_\alpha \psi(\alpha_1) - d_\beta \psi(\beta_1) + (d_\alpha + d_\beta)\psi(\alpha_1 + \beta_1)$$

where $d_\alpha = \alpha_2 - \alpha_1$ and $\psi(v) = \frac{\Gamma'(v)}{\Gamma(v)}$ is the digamma function. For each choice of $P$ and $Q$, a 1-dimensional and a 10-dimensional experiment was performed. In the *1-d experiment*, samples of $n = 100$ values, $X_{100}$ and $Y_{100}$, were each drawn i.i.d. from $P$ and $Q$ respectively. In the *10-d experiment*, each $x \in X_{100} \subset \mathbb{R}^{10}$ and $y \in Y_{100} \subset \mathbb{R}^{10}$ was drawn i.i.d. from the respective product distributions $P \times \prod_{i=1}^{9} N(0, 0.01)$ and $Q \times \prod_{i=1}^{9} N(0, 0.01)$. This gives samples from two distributions embedded in a 10-dimensional space where all but one of the dimensions is zero mean Gaussian noise. The KL divergences for the 10-d product distributions for each choice of $P$ and $Q$ are the same as for the 1-d case, that is, $KL(P, Q)$. The specific $P$ and $Q$ in the experiments were chosen to give a range of different KL divergence values and explore a few different pairings of distributional shapes.

### 3.2. Results

Table 1 summarises the application of all five estimators over 250 runs of the 1-d (odd rows) and 10-d experiments (even rows) for various choices of $P$ and $Q$ shown in the first column. The pairs of rows are ordered in increasing value of true KL divergence (shown in the second column) and is the same for both rows. The table lists the divergence estimates averaged over the different runs as well as the empirical Mean Squared Error (MSE). The bold values for the MSE correspond to the lowest amongst the different estimators. Where the MSE of our estimator or that of NWJ is strictly lower than the other, we have italicised the MSE. The last three columns are in grey as they are not the main point of the experiments.

### 3.3. Discussion

For the most part our estimator performs better in terms of MSE than that of the NWJ. When the true divergence is large, the difference is especially pronounced. This is unsurprising as our estimator is based on a tighter bound for the divergence as Theorem 1 shows. For small divergences the difference is smaller since roughly speaking, $0 \le \text{NWJ} \le \text{Ours} \le KL(P, Q)$. Thus, for small divergences the estimators must necessarily return similar values.

In contrast, the nearest neighbour-based methods (WKV and $(f, l)$) behave very differently to variational estimators. In general, their bias is significantly lower than both variational methods when the real divergence is large. This is a natural conclusion, since the variational methods presented here are intrinsically lower bounds of the real divergence. Finally, we note that uLSIF does not perform as well as the other methods. This is to be expected as uLSIF is primarily designed for density ratio estimation while the rest of the methods are derived specifically for divergence estimation.

---

[4] `http://sugiyama-www.cs.titech.ac.jp/~sugi/software/uLSIF`

[5] `http://www.tsc.uc3m.es/~dggarcia/code.html`

[6] All code has been submitted as supplementary material.

*Table 1.* Summary of results for KL divergence estimation by the restricted variational estimator (Ours), Nguyen et al.'s (NWJ), Wang et al.'s (WKV), Kanamori et al (uLSIF) and García et al (($f, l$))

| Distributions | KL | Estimates | | | | | MSE | | | | |
|---|---|---|---|---|---|---|---|---|---|---|---|
| | | Ours | NWJ | WKV | uLSIF | ($f, l$) | Ours | NWJ | WKV | uLSIF | ($f, l$) |
| B(2,2) vs B(4,4) | 0.183 | 0.15 | 0.13 | 0.13 | 0.27 | 0.33 | **0.006** | **0.006** | 0.047 | 0.111 | 0.104 |
| | | 0.15 | 0.13 | 0.47 | 0.347 | 0.28 | **0.006** | **0.006** | 0.325 | 0.254 | 0.081 |
| B(1,1) vs B(2,2) | 0.208 | 0.17 | 0.16 | 0.18 | 0.34 | 0.18 | **0.008** | **0.008** | 0.050 | 0.263 | 0.032 |
| | | 0.18 | 0.17 | 0.41 | 0.44 | 0.31 | **0.007** | **0.007** | 0.213 | 0.383 | 0.086 |
| B(14,14) vs B(1,1) | 0.959 | 0.83 | 0.79 | 0.97 | 0.587 | 1.386 | *0.036* | 0.046 | 0.062 | 0.367 | 0.365 |
| | | 0.82 | 0.79 | 1.24 | 0.61 | 1.42 | *0.040* | 0.051 | 0.226 | 0.195 | 0.451 |
| B(1,2) vs B(2,1) | 1.000 | 0.87 | 0.86 | 0.95 | 1.26 | 1.18 | 0.074 | ***0.072*** | 0.120 | 0.581 | 0.113 |
| | | 0.86 | 0.84 | 1.92 | 0.73 | 1.48 | **0.06** | **0.06** | 1.648 | 0.471 | 0.397 |
| B(1,1) vs B(5,5) | 1.554 | 0.77 | 0.63 | 0.86 | 1.12 | 1.47 | *0.647* | 0.859 | 0.563 | 0.592 | **0.251** |
| | | 0.72 | 0.59 | 2.81 | 0.98 | 1.18 | *0.737* | 0.954 | 2.322 | 1.088 | **0.323** |
| B(1,4) vs B(3,1) | 3.704 | 2.79 | 2.68 | 3.03 | 4.67 | 3.74 | *0.985* | 1.153 | 0.729 | 9.229 | **0.159** |
| | | 2.19 | 2.11 | 8.27 | 0.83 | 3.86 | *2.361* | 2.653 | 28.685 | 11.48 | **0.174** |
| B(1,1) vs B(10,10) | 4.264 | 1.40 | 0.99 | 1.47 | 1.87 | 2.33 | *8.257* | 10.741 | 7.869 | 7.265 | **3.994** |
| | | 1.20 | 0.83 | 5.61 | 1.68 | 1.955 | *9.434* | 11.829 | 3.249 | 7.556 | 5.511 |

On the 10-d experiments, the MSE performance of the WKV estimator is typically much worse than the rest of methods. It consistently over-estimates the true KL divergence and, for $B(1, 4)$ vs $B(3, 1)$, drastically overshoots it, resulting in an order of magnitude larger MSE than the other estimators. One likely explanation of this poor performance of WKV on the higher dimensional problems is that its estimated values scale with the dimension of the data. This scaling occurs even if the two distributions differ only on a low dimensional manifold, as they do in the 10-d experiments. The success of this estimator in the $B(1, 1)$ vs $B(10, 10)$ experiment is likely a coincidence. All the estimators underestimate this divergence in the 1-d case and we expect that the scaling of WKV with the dimension has pushed its estimate up to the true KL. The ($f, l$) estimator, although also based on nearest neighbour techniques, does not suffer from this problem since it does not present a explicit dependance of the estimated divergence with respect to the ambient dimension.

In light of these observations, we offer some guidelines as to which estimator to use if one has some prior knowledge or suspicion about the data. Use our method when suspecting a low divergence; use ($f, l$) for high divergence. We also recommend using variational methods over nearest neighbour estimators for hypothesis testing if false positives are a concern since the variational methods are much more likely to consistently underestimate the true divergence. It is important however to note that if running time is an issue then WKV becomes a very attractive option. There are many fast approximate nearest neighbour algorithms resulting in fast estimation of the WKV statistic.

## 4. Summary and Discussion

We have shown how tighter variational representations for $f$-divergences can be derived by restricting the effective domain of the divergence functional to the set of probability measures. Since many works in the literature are based on variational representations, this tighter version presents many potential applications. As an example of this, a dual program for $f$-divergence estimators based on this tighter representation was derived for density ratios within an RKHS $\mathcal{H}$ and arbitrary convex regularizers. This tightened and extended the M2 estimator proposed in (Nguyen et al., 2010) and we demonstrated empirically the benefits of our analysis. We also gave a novel interpretation of the dual program in terms of MMD which showed that our estimator can be seen to find an approximation $\hat{r} \in \mathcal{H}$ of the density ratio that attempts to simultaneously minimises MMD between $P_n$ and $\hat{r}Q_n$ and the empirical $f$-divergence $\mathbb{E}_{Q_n}[f(\hat{r})]$. This second minimisation can be seen as a generalised maximum entropy regularisation. We have also provided a comparison to other state of the art estimators. We concluded that variational methods are good for settings in which a low divergence is suspected or in scenarios where overestimation is detrimental.

As future work we intend to investigate the impact of this tightened representation on other divergence estimators based on the looser representation such as (Kanamori et al., 2011), as well as to areas other than $f$-divergence estimation (hypothesis testing and statistical inference). We also plan to find general conditions under which consistency of our family of estimators holds. The work of (Nguyen et al., 2010) has already paved the way for this investigation. Failing that, the very general consistency results of (Altun & Smola, 2006) for single sample divergence estimation may also be amenable to the analysis of our estimator. The performance of our estimator on distributions on low dimensional manifolds suggests that it would be worth testing on domains involving audio or images. It would also be interesting to apply our method for estimating divergences other than KL. For instance we could study $\alpha$-divergence estimation as in (Poczos & Schneider, 2011).